# A Universal Approximation Theorem for Mixture of Experts Models

Hien D. Nguyen[1], Luke R. Lloyd-Jones[2], and Geoffrey J. McLachlan[1*]

February 11, 2016

**Abstract**

The mixture of experts (MoE) model is a popular neural network architecture for nonlinear regression and classification. The class of MoE mean functions is known to be uniformly convergent to any unknown target function, assuming that the target function is from Sobolev space that is sufficiently differentiable and that the domain of estimation is a compact unit hypercube. We provide an alternative result, which shows that the class of MoE mean functions is dense in the class of all continuous functions over arbitrary compact domains of estimation. Our result can be viewed as a universal approximation theorem for MoE models.

## 1 Introduction

The mixture of experts (MoE) model is a neural network architecture for nonlinear regression and classification. The MoE model was introduced in [6, 10]; reviews of research in MoE models can be found in Section 5.13 of [12] and [17]. Examples of recent research in MoE models include [1] and [14], where MoE models are used for curve classification and robust estimation, respectively.

Let $Y \in \mathbb{R}$ be a random variable and let $\boldsymbol{x} \in \mathbb{X} \subset \mathbb{R}^p$ be a $p$-dimensional vector. Let the conditional probability density function of $Y$ given $\boldsymbol{x}$ be

$$f_g(y|\boldsymbol{x}) = \sum_{i=1}^{g} \pi_i(\boldsymbol{x};\boldsymbol{\psi}_g)\, h\left(y;\gamma_i + \boldsymbol{x}^T\boldsymbol{\delta}_i, \boldsymbol{\nu}_i\right), \qquad (1)$$

where

$$\pi_i(\boldsymbol{x};\boldsymbol{\psi}_g) = \frac{\exp\left(\alpha_i + \boldsymbol{x}^T\boldsymbol{\beta}_i\right)}{\sum_{j=1}^{g}\exp\left(\alpha_j + \boldsymbol{x}^T\boldsymbol{\beta}_j\right)},$$

$y$ is a realization of $Y$, and $h(y;\mu,\boldsymbol{\xi})$ is a univariate component probability density function (PDF) (in $Y$) with mean $\mu$ and nuisance parameter $\boldsymbol{\xi}$. Here $\alpha_i, \gamma_i \in$

*[1]School of Mathematics and Physics, University of Queensland (Email: h.nguyen7@uq.edu.au; g.mclachlan@uq.edu.au). [2]Queensland Brain Institute, University of Queensland (Email: l.lloydjones@uq.edu.au).

$\mathbb{R}$, $\boldsymbol{\beta}_i, \boldsymbol{\delta}_i \in \mathbb{R}^p$, and $\boldsymbol{\nu}_i \in \mathbb{R}^q$ for each $i = 1, ..., g$, and $\boldsymbol{\psi}_g^T = (\alpha_1, \boldsymbol{\beta}_1, ..., \alpha_g, \boldsymbol{\beta}_g)$. We say that (1) is a $g$-component MoE with mean function

$$\mu_g(\boldsymbol{x}; \boldsymbol{\theta}_g) = \sum_{i=1}^{g} \pi_i(\boldsymbol{x}; \boldsymbol{\psi}_g) \left(\gamma_i + \boldsymbol{x}^T \boldsymbol{\delta}_i\right), \tag{2}$$

where $\boldsymbol{\theta}_g^T = (\boldsymbol{\psi}_g, \gamma_1, \boldsymbol{\delta}_1, ..., \boldsymbol{\psi}_g, \gamma_g, \boldsymbol{\delta}_g)$ is the function's parameter vector. The superscript $T$ notation indicates matrix transposition.

In [18], it was shown that there exists a sequence of functions $\mu_g(\boldsymbol{x}; \boldsymbol{\theta}_g)$ that converges uniformly to any target function $m(\boldsymbol{x})$, in the index $g$, assuming that $m(\boldsymbol{x})$ belongs to a Sobolev class of functions and $\mathbb{X}$ is a closed unit hypercube. The result from [18] was generalized to nonlinear mappings of the expressions $\gamma_i + \boldsymbol{x}^T \boldsymbol{\delta}_i$ in the component PDFs of the MoE, in [8]. The result from [8] was expanded upon in [7] where it was shown that there exists a sequence of conditional PDFs $f_g(y|\boldsymbol{x})$ that converges in Kullback-Leibler divergence to any target conditional PDF $f(y|\boldsymbol{x})$, in $g$, assuming that $f(y|\boldsymbol{x})$ belongs to the one-parameter exponential family of density functions. Convergence results for MoE models with polynomial component PDF mean functions were then obtained in [13]. We note that the target mean function $m(\boldsymbol{x})$ is assumed to belong to a Sobolev class of functions in each of [8, 7, 13], as they are in [18].

Define the class of all mean functions of form (2) as

$$\mathbb{M} = \left\{\mu_g(\boldsymbol{x}; \boldsymbol{\theta}_g) | g \in \mathbb{N}, \boldsymbol{\theta}_g \in \mathbb{R}^{g(2p+2)}\right\}$$

and let $\mathbb{C}(\mathbb{X})$ be the class of continuous functions on the domain $\mathbb{X}$. In this article, we prove that $\mathbb{M}$ is dense in the set $\mathbb{C}(\mathbb{X})$ under the assumption that $\mathbb{X}$ is compact. Our result is obtained via the Stone-Weierstrass Theorem [16]; see also [2] for a discussion in the context of neural networks. Our result is a universal approximation theorem for MoE mean functions in the style of [3]. It can be seen as an alternative to the result of [18] that makes no assumptions regarding the shape of the domain $\mathbb{X}$, nor imposes any differentiability restrictions on the class of target function.

The article proceeds as follows. In Section II, we present the Stone-Weierstrass Theorem. In Section III, we prove that $\mathbb{M}$ fulfills the conditions of the Stone-Weierstrass Theorem and thus is dense in the class $\mathbb{C}(\mathbb{X})$. Conclusions are drawn in Section IV.

# 2 Stone-Weierstrass Theorem

Following the presentation of [2], the Stone-Weierstrass Theorem can be phrased as follows.

**Theorem 1.** *Let $\mathbb{X} \subset \mathbb{R}^p$ be a compact set and let $\mathbb{U}$ be a set of continuous real-valued functions on $\mathbb{X}$. Assume that*

**(i)** *the constant function $u(\boldsymbol{x}) = 1$ is in $\mathbb{U}$.*

**(ii)** *for any two points* $\boldsymbol{x}_1, \boldsymbol{x}_2 \in \mathbb{X}$ *such that* $\boldsymbol{x}_1 \neq \boldsymbol{x}_2$*, there exists a function* $u \in \mathbb{U}$ *such that* $u(\boldsymbol{x}_1) \neq u(\boldsymbol{x}_2)$.

**(iii)** *if* $a \in \mathbb{R}$ *and* $u \in \mathbb{U}$*, then* $au \in \mathbb{U}$.

**(iv)** *if* $u, v \in \mathbb{U}$*, then* $uv \in \mathbb{U}$.

**(v)** *if* $u, v \in \mathbb{U}$*, then* $u + v \in \mathbb{U}$.

*If Assumptions (i)–(v) are true, then* $\mathbb{U}$ *is dense in* $\mathbb{C}(\mathbb{X})$*. In other words, for any* $\epsilon > 0$ *and any* $v \in \mathbb{C}(\mathbb{X})$*, there exists a* $u \in \mathbb{U}$ *such that* $\sup_{\boldsymbol{x} \in \mathbb{X}} |u(\boldsymbol{x}) - v(\boldsymbol{x})| < \epsilon$.

For completeness, we note that $\mathbb{X}$ is compact if and only if it is bounded and closed in the standard Euclidean space. See Chapter 2 of [4] for detailed discussions regarding both set compactness and the Stone-Weierstrass Theorem.

# 3 Main Result

Define $\mathbf{0}$ to be a vector of zeros of an appropriate dimensionality. In order to facilitate the proof, let

$$\mathbb{H} = \left\{ \eta_g(\boldsymbol{x}; \boldsymbol{\omega}_g) \,|\, g \in \mathbb{N}, \boldsymbol{\omega}_g \in \mathbb{R}^{g(p+2)} \right\},$$

where

$$\eta_g(\boldsymbol{x}; \boldsymbol{\omega}_g) = \sum_{i=1}^{g} \gamma_i \pi_i(\boldsymbol{x}; \boldsymbol{\psi}_g)$$

and $\boldsymbol{\omega}_g^T = (\boldsymbol{\psi}_1, \gamma_1, ..., \boldsymbol{\psi}_g, \gamma_g)$ is the function's parameter vector. Note that

$$\mu_g\left(\boldsymbol{x}; \tilde{\boldsymbol{\theta}}_g\right) = \eta_g(\boldsymbol{x}; \boldsymbol{\omega}_g),$$

if $\tilde{\boldsymbol{\theta}}_g^T = (\boldsymbol{\psi}_g, \gamma_1, \mathbf{0}, ..., \boldsymbol{\psi}_g, \gamma_g, \mathbf{0})$. Thus, it can be seen that $\mathbb{H} \subset \mathbb{M}$. If we can show that for all $m \in \mathbb{C}(\mathbb{X})$, there exists an $\eta \in \mathbb{H}$ such that $\sup_{\boldsymbol{x} \in \mathbb{X}} |\eta(\boldsymbol{x}) - m(\boldsymbol{x})| < \epsilon$, for any $\epsilon > 0$, then we can choose $\mu \in \mathbb{M}$ such that $\sup_{\boldsymbol{x} \in \mathbb{X}} |\mu(\boldsymbol{x}) - m(\boldsymbol{x})| < \epsilon$, by setting $\mu = \eta \in \mathbb{M}$.

We now proceed to prove that $\mathbb{H}$ is dense in $\mathbb{C}(\mathbb{X})$. We obtain the result via the following five lemmas, which show that $\mathbb{H}$ fulfills Assumptions (i)–(v) of Theorem 1. Since $\mathbb{H} \subset \mathbb{M}$, we obtain the main result of the article as a corollary.

**Lemma 2.** *The constant function* $\eta(\boldsymbol{x}) = 1$ *is in* $\mathbb{H}$.

*Proof.* Let $g = 1$ and $\tilde{\boldsymbol{\omega}}_1^T = (\boldsymbol{\psi}_1, 1)$. Set $\eta(\boldsymbol{x}) = \eta_1(\boldsymbol{x}; \tilde{\boldsymbol{\omega}}_1)$. For any choice of $\boldsymbol{\psi}_1$, $\eta(\boldsymbol{x}) = 1$. We obtain the desired result by noting that $\eta_1(\boldsymbol{x}; \tilde{\boldsymbol{\omega}}_1) \in \mathbb{H}$. □

**Lemma 3.** *For any two points* $\boldsymbol{x}_1, \boldsymbol{x}_2 \in \mathbb{X}$ *such that* $\boldsymbol{x}_1 \neq \boldsymbol{x}_2$*, there exists a function* $\eta \in \mathbb{H}$ *such that* $\eta(\boldsymbol{x}_1) \neq \eta(\boldsymbol{x}_2)$.

*Proof.* Let $g = 2$ and $\bar{\boldsymbol{\omega}}_2^T = (\boldsymbol{\psi}_1, 0, \mathbf{0}, \gamma_2)$, where $\gamma_2 \neq 0$. Set $\eta(\boldsymbol{x}) = \eta_2(\boldsymbol{x}; \bar{\boldsymbol{\omega}}_2)$ and assume that $\boldsymbol{x}_j^T = (x_{j1}, ..., x_{jp})$ for $j = 1, 2$, such that $\boldsymbol{x}_1 \neq \boldsymbol{x}_2$. Let $\eta(\boldsymbol{x}_1) \neq \eta(\boldsymbol{x}_2)$; this is equivalent to

$$\begin{aligned} \eta_2(\boldsymbol{x}_1; \bar{\boldsymbol{\omega}}_2) &\neq \eta_2(\boldsymbol{x}_2; \bar{\boldsymbol{\omega}}_2) \\ \frac{\gamma_2}{1 + \exp\left(\alpha_1 + \boldsymbol{x}_1^T \boldsymbol{\beta}_1\right)} &\neq \frac{\gamma_2}{1 + \exp\left(\alpha_1 + \boldsymbol{x}_2^T \boldsymbol{\beta}_1\right)} \end{aligned}$$

by substitution, and reduces to

$$(\boldsymbol{x}_1 - \boldsymbol{x}_2)^T \boldsymbol{\beta}_1 \neq 0. \tag{3}$$

Equation (3) is violated if either $\boldsymbol{x}_1 = \boldsymbol{x}_2$, which causes a contradiction, or if $\boldsymbol{\beta}_1^T = (\beta_1, ..., \beta_p)$ is such that $\beta_k = 0$ whenever $x_{1k} \neq x_{2k}$, for $k = 1, ..., p$. To avoid violation of (3), we can set $\beta_k \neq 0$ for all $k$.

Thus, let $\tilde{\boldsymbol{\omega}}_2^T = \left(\tilde{\boldsymbol{\psi}}_1, 0, \mathbf{0}, \gamma_2\right)$ and $\eta(\boldsymbol{x}) = \eta_2(\boldsymbol{x}; \tilde{\boldsymbol{\omega}}_2)$, where $\tilde{\boldsymbol{\psi}}_1 \in \{\psi \in \mathbb{R} | \psi \neq 0\}^{p+1}$ and $\gamma_2 \neq 0$. If $\boldsymbol{x}_1 \neq \boldsymbol{x}_2$, then $\eta(\boldsymbol{x}_1) \neq \eta(\boldsymbol{x}_2)$. We obtain the desired result by noting that $\eta_2(\boldsymbol{x}; \tilde{\boldsymbol{\omega}}_2) \in \mathbb{H}$. □

**Lemma 4.** *If $a \in \mathbb{R}$ and $\eta \in \mathbb{H}$, then $a\eta \in \mathbb{H}$.*

*Proof.* Let $a \in \mathbb{R}$ and $\eta(\boldsymbol{x}) = \eta_g(\boldsymbol{x}; \boldsymbol{\omega}_g)$. We can write

$$\begin{aligned} a\eta(\boldsymbol{x}) &= a \sum_{i=1}^{g} \gamma_i \pi_i(\boldsymbol{x}; \boldsymbol{\psi}_g) \\ &= \sum_{i=1}^{g} (a\gamma_i) \pi_i(\boldsymbol{x}; \boldsymbol{\psi}_g) \\ &= \sum_{i=1}^{g} \tilde{\gamma}_i \pi_i(\boldsymbol{x}; \boldsymbol{\psi}_g), \end{aligned}$$

where $\tilde{\gamma}_i = a\gamma_i$ for $i = 1, ..., g$. Thus, $a\eta(\boldsymbol{x}) = \eta_g(\boldsymbol{x}; \tilde{\boldsymbol{\omega}}_g)$, where $\tilde{\boldsymbol{\omega}}_g^T = (\boldsymbol{\psi}_1, \tilde{\gamma}_1, ..., \boldsymbol{\psi}_g, \tilde{\gamma}_g)$. We obtain the desired result by noting that $\eta_g(\boldsymbol{x}; \tilde{\boldsymbol{\omega}}_g) \in \mathbb{H}$. □

**Lemma 5.** *If $\eta, \lambda \in \mathbb{H}$, then $\eta\lambda \in \mathbb{H}$.*

*Proof.* Let $g, m \in \mathbb{N}$, $\boldsymbol{\omega}_g^{[\eta]T} = \left(\boldsymbol{\psi}_1^{[\eta]}, \gamma_1^{[\eta]}, ..., \boldsymbol{\psi}_g^{[\eta]}, \gamma_g^{[\eta]}\right)$, and $\boldsymbol{\omega}_m^{[\lambda]T} = \left(\boldsymbol{\psi}_1^{[\lambda]}, \gamma_1^{[\lambda]}, ..., \boldsymbol{\psi}_g^{[\lambda]}, \gamma_g^{[\lambda]}\right)$, and set $\eta(\boldsymbol{x}) = \eta_g\left(\boldsymbol{x}; \boldsymbol{\omega}_g^{[\eta]}\right)$ and $\lambda(\boldsymbol{x}) = \eta_m\left(\boldsymbol{x}; \boldsymbol{\omega}_m^{[\lambda]}\right)$. Here, the superscripts $[\eta]$ and $[\lambda]$ denote the parameter components belonging to the functions $\eta$ and $\lambda$, respectively. We can write

$$\begin{aligned} \eta(\boldsymbol{x}) \lambda(\boldsymbol{x}) &= \sum_{i=1}^{g} \gamma_i^{[\eta]} \pi_i\left(\boldsymbol{x}; \boldsymbol{\psi}_g^{[\eta]}\right) \sum_{j=1}^{m} \gamma_j^{[\lambda]} \pi_j\left(\boldsymbol{x}; \boldsymbol{\psi}_m^{[\lambda]}\right) \\ &= \sum_{i=1}^{g} \sum_{j=1}^{m} \gamma_i^{[\eta]} \gamma_j^{[\lambda]} \pi_i\left(\boldsymbol{x}; \boldsymbol{\psi}_g^{[\eta]}\right) \pi_j\left(\boldsymbol{x}; \boldsymbol{\psi}_m^{[\lambda]}\right). \end{aligned} \tag{4}$$

To simplify (4), for each $i=1,...,g$ and $j=1,...,m$, we can write

$$\begin{aligned}
& \pi_i\left(\boldsymbol{x};\boldsymbol{\psi}_g^{[\eta]}\right)\pi_j\left(\boldsymbol{x};\boldsymbol{\psi}_m^{[\lambda]}\right) \qquad (5)\\
= & \frac{\exp\left(\alpha_i^{[\eta]}+\boldsymbol{x}^T\boldsymbol{\beta}_i^{[\eta]}\right)\exp\left(\alpha_j^{[\lambda]}+\boldsymbol{x}^T\boldsymbol{\beta}_j^{[\lambda]}\right)}{\sum_{k=1}^{g}\exp\left(\alpha_k^{[\eta]}+\boldsymbol{x}^T\boldsymbol{\beta}_k^{[\eta]}\right)\sum_{l=1}^{m}\exp\left(\alpha_l^{[\lambda]}+\boldsymbol{x}^T\boldsymbol{\beta}_l^{[\lambda]}\right)}\\
= & \frac{\exp\left(\left[\alpha_i^{[\eta]}+\alpha_j^{[\lambda]}\right]+\boldsymbol{x}^T\left[\boldsymbol{\beta}_i^{[\eta]}+\boldsymbol{\beta}_j^{[\lambda]}\right]\right)}{\sum_{k=1}^{g}\sum_{l=1}^{m}\exp\left(\left[\alpha_k^{[\eta]}+\alpha_l^{[\lambda]}\right]+\boldsymbol{x}^T\left[\boldsymbol{\beta}_k^{[\eta]}+\boldsymbol{\beta}_l^{[\lambda]}\right]\right)}.
\end{aligned}$$

Upon performing the mapping from Table I-A, we can write the final line of (5) as $\pi_k\left(\boldsymbol{x};\tilde{\boldsymbol{\psi}}_{gm}\right)$, where $\tilde{\boldsymbol{\psi}}_k^T=\left(\tilde{\alpha}_k,\tilde{\boldsymbol{\beta}}_k\right)$ for $k=1,...,gm$. Furthermore, via the mapping from Table I-A, (4) can be simplified to

$$\begin{aligned}
\eta\left(\boldsymbol{x}\right)\lambda\left(\boldsymbol{x}\right) &= \sum_{k=1}^{gm}\tilde{\gamma}_k\pi_k\left(\boldsymbol{x};\tilde{\boldsymbol{\psi}}_{gm}\right)\\
&= \eta_{gm}\left(\boldsymbol{x};\tilde{\boldsymbol{\omega}}_{gm}\right),
\end{aligned}$$

where $\tilde{\boldsymbol{\omega}}_{gm}^T=\left(\tilde{\boldsymbol{\psi}}_1,\tilde{\gamma}_1,...,\tilde{\boldsymbol{\psi}}_{gm},\tilde{\gamma}_{gm}\right)$. We obtain the desired result by noting that $\eta_{gm}\left(\boldsymbol{x};\tilde{\boldsymbol{\omega}}_{gm}\right)\in\mathbb{H}$.

□

**Lemma 6.** *If $\eta,\lambda\in\mathbb{H}$, then $\eta+\lambda\in\mathbb{H}$.*

*Proof.* Let $g,m\in\mathbb{N}$, $\boldsymbol{\omega}_g^{[\eta]T}=\left(\boldsymbol{\psi}_1^{[\eta]},\gamma_1^{[\eta]},...,\boldsymbol{\psi}_g^{[\eta]},\gamma_g^{[\eta]}\right)$, and $\boldsymbol{\omega}_m^{[\lambda]T}=\left(\boldsymbol{\psi}_1^{[\lambda]},\gamma_1^{[\lambda]},...,\boldsymbol{\psi}_g^{[\lambda]},\gamma_g^{[\lambda]}\right)$, and set $\eta\left(\boldsymbol{x}\right)=\eta_g\left(\boldsymbol{x};\boldsymbol{\omega}_g^{[\eta]}\right)$ and $\lambda\left(\boldsymbol{x}\right)=\eta_m\left(\boldsymbol{x};\boldsymbol{\omega}_m^{[\lambda]}\right)$. Here, the superscripts $[\eta]$ and $[\lambda]$ denote the parameter components belonging to the functions $\eta$ and $\lambda$, respectively. We can write

$$\begin{aligned}
& \eta\left(\boldsymbol{x}\right)+\lambda\left(\boldsymbol{x}\right) \qquad (6)\\
= & \sum_{i=1}^{g}\gamma_i^{[\eta]}\pi_i\left(\boldsymbol{x};\boldsymbol{\psi}_g^{[\eta]}\right)+\sum_{j=1}^{m}\gamma_j^{[\lambda]}\pi_j\left(\boldsymbol{x};\boldsymbol{\psi}_m^{[\lambda]}\right)\\
= & \frac{\sum_{i=1}^{g}\gamma_i^{[\eta]}\exp\left(\alpha_i^{[\eta]}+\boldsymbol{x}^T\boldsymbol{\beta}_i^{[\eta]}\right)}{\sum_{k=1}^{g}\exp\left(\alpha_k^{[\eta]}+\boldsymbol{x}^T\boldsymbol{\beta}_k^{[\eta]}\right)}\\
& +\frac{\sum_{j=1}^{m}\gamma_j^{[\lambda]}\exp\left(\alpha_j^{[\lambda]}+\boldsymbol{x}^T\boldsymbol{\beta}_j^{[\lambda]}\right)}{\sum_{l=1}^{m}\exp\left(\alpha_l^{[\lambda]}+\boldsymbol{x}^T\boldsymbol{\beta}_l^{[\lambda]}\right)}\\
= & \frac{\sum_{i=1}^{g}\gamma_i^{[\eta]}\exp\left(\alpha_i^{[\eta]}+\boldsymbol{x}^T\boldsymbol{\beta}_i^{[\eta]}\right)\sum_{l=1}^{m}\exp\left(\alpha_j^{[\lambda]}+\boldsymbol{x}^T\boldsymbol{\beta}_j^{[\lambda]}\right)}{\sum_{k=1}^{g}\exp\left(\alpha_k^{[\eta]}+\boldsymbol{x}^T\boldsymbol{\beta}_k^{[\eta]}\right)\sum_{l=1}^{m}\exp\left(\alpha_l^{[\lambda]}+\boldsymbol{x}^T\boldsymbol{\beta}_l^{[\lambda]}\right)}
\end{aligned}$$

Table 1: Mapping of parameter components for Lemmas 5 and 6.

A

| | | | | | | | | |
|---|---|---|---|---|---|---|---|---|
| $\alpha_1^{[\eta]} + \alpha_1^{[\lambda]}$ | $\longrightarrow$ | $\tilde{\alpha}_1$ | $\boldsymbol{\beta}_1^{[\eta]} + \boldsymbol{\beta}_1^{[\lambda]}$ | $\longrightarrow$ | $\tilde{\boldsymbol{\beta}}_1$ | $\gamma_1^{[\eta]}\gamma_1^{[\lambda]}$ | $\longrightarrow$ | $\tilde{\gamma}_1$ |
| $\alpha_1^{[\eta]} + \alpha_2^{[\lambda]}$ | $\longrightarrow$ | $\tilde{\alpha}_2$ | $\boldsymbol{\beta}_1^{[\eta]} + \boldsymbol{\beta}_2^{[\lambda]}$ | $\longrightarrow$ | $\tilde{\boldsymbol{\beta}}_2$ | $\gamma_1^{[\eta]}\gamma_2^{[\lambda]}$ | $\longrightarrow$ | $\tilde{\gamma}_2$ |
| $\vdots$ | | | $\vdots$ | | | $\vdots$ | | |
| $\alpha_1^{[\eta]} + \alpha_m^{[\lambda]}$ | $\longrightarrow$ | $\tilde{\alpha}_{m+1}$ | $\boldsymbol{\beta}_1^{[\eta]} + \boldsymbol{\beta}_m^{[\lambda]}$ | $\longrightarrow$ | $\tilde{\boldsymbol{\beta}}_{m+1}$ | $\gamma_1^{[\eta]}\gamma_m^{[\lambda]}$ | $\longrightarrow$ | $\tilde{\gamma}_{m+1}$ |
| $\alpha_2^{[\eta]} + \alpha_1^{[\lambda]}$ | $\longrightarrow$ | $\tilde{\alpha}_{m+2}$ | $\boldsymbol{\beta}_2^{[\eta]} + \boldsymbol{\beta}_1^{[\lambda]}$ | $\longrightarrow$ | $\tilde{\boldsymbol{\beta}}_{m+2}$ | $\gamma_2^{[\eta]}\gamma_1^{[\lambda]}$ | $\longrightarrow$ | $\tilde{\gamma}_{m+2}$ |
| $\vdots$ | | | $\vdots$ | | | $\vdots$ | | |
| $\alpha_k^{[\eta]} + \alpha_m^{[\lambda]}$ | $\longrightarrow$ | $\tilde{\alpha}_{km+1}$ | $\boldsymbol{\beta}_k^{[\eta]} + \boldsymbol{\beta}_m^{[\lambda]}$ | $\longrightarrow$ | $\tilde{\boldsymbol{\beta}}_{km+1}$ | $\gamma_k^{[\eta]}\gamma_m^{[\lambda]}$ | $\longrightarrow$ | $\tilde{\gamma}_{km+1}$ |
| $\alpha_{k+1}^{[\eta]} + \alpha_1^{[\lambda]}$ | $\longrightarrow$ | $\tilde{\alpha}_{km+2}$ | $\boldsymbol{\beta}_{k+1}^{[\eta]} + \boldsymbol{\beta}_1^{[\lambda]}$ | $\longrightarrow$ | $\tilde{\boldsymbol{\beta}}_{km+2}$ | $\gamma_{k+1}^{[\eta]}\gamma_1^{[\lambda]}$ | $\longrightarrow$ | $\tilde{\gamma}_{km+2}$ |
| $\vdots$ | | | $\vdots$ | | | $\vdots$ | | |
| $\alpha_g^{[\eta]} + \alpha_{m-1}^{[\lambda]}$ | $\longrightarrow$ | $\tilde{\alpha}_{gm-1}$ | $\boldsymbol{\beta}_g^{[\eta]} + \boldsymbol{\beta}_{m-1}^{[\lambda]}$ | $\longrightarrow$ | $\tilde{\boldsymbol{\beta}}_{gm-1}$ | $\gamma_g^{[\eta]}\gamma_{m-1}^{[\lambda]}$ | $\longrightarrow$ | $\tilde{\gamma}_{gm-1}$ |
| $\alpha_g^{[\eta]} + \alpha_m^{[\lambda]}$ | $\longrightarrow$ | $\tilde{\alpha}_{gm}$ | $\boldsymbol{\beta}_g^{[\eta]} + \boldsymbol{\beta}_m^{[\lambda]}$ | $\longrightarrow$ | $\tilde{\boldsymbol{\beta}}_{gm}$ | $\gamma_g^{[\eta]}\gamma_m^{[\lambda]}$ | $\longrightarrow$ | $\tilde{\gamma}_{gm}$ |

B

| | | | | | | | | |
|---|---|---|---|---|---|---|---|---|
| $\alpha_1^{[\eta]} + \alpha_1^{[\lambda]}$ | $\longrightarrow$ | $\tilde{\alpha}_1$ | $\boldsymbol{\beta}_1^{[\eta]} + \boldsymbol{\beta}_1^{[\lambda]}$ | $\longrightarrow$ | $\tilde{\boldsymbol{\beta}}_1$ | $\gamma_1^{[\eta]} + \gamma_1^{[\lambda]}$ | $\longrightarrow$ | $\tilde{\gamma}_1$ |
| $\alpha_1^{[\eta]} + \alpha_2^{[\lambda]}$ | $\longrightarrow$ | $\tilde{\alpha}_2$ | $\boldsymbol{\beta}_1^{[\eta]} + \boldsymbol{\beta}_2^{[\lambda]}$ | $\longrightarrow$ | $\tilde{\boldsymbol{\beta}}_2$ | $\gamma_1^{[\eta]} + \gamma_2^{[\lambda]}$ | $\longrightarrow$ | $\tilde{\gamma}_2$ |
| $\vdots$ | | | $\vdots$ | | | $\vdots$ | | |
| $\alpha_1^{[\eta]} + \alpha_m^{[\lambda]}$ | $\longrightarrow$ | $\tilde{\alpha}_{m+1}$ | $\boldsymbol{\beta}_1^{[\eta]} + \boldsymbol{\beta}_m^{[\lambda]}$ | $\longrightarrow$ | $\tilde{\boldsymbol{\beta}}_{m+1}$ | $\gamma_1^{[\eta]} + \gamma_m^{[\lambda]}$ | $\longrightarrow$ | $\tilde{\gamma}_{m+1}$ |
| $\alpha_2^{[\eta]} + \alpha_1^{[\lambda]}$ | $\longrightarrow$ | $\tilde{\alpha}_{m+2}$ | $\boldsymbol{\beta}_2^{[\eta]} + \boldsymbol{\beta}_1^{[\lambda]}$ | $\longrightarrow$ | $\tilde{\boldsymbol{\beta}}_{m+2}$ | $\gamma_2^{[\eta]} + \gamma_1^{[\lambda]}$ | $\longrightarrow$ | $\tilde{\gamma}_{m+2}$ |
| $\vdots$ | | | $\vdots$ | | | $\vdots$ | | |
| $\alpha_k^{[\eta]} + \alpha_m^{[\lambda]}$ | $\longrightarrow$ | $\tilde{\alpha}_{km+1}$ | $\boldsymbol{\beta}_k^{[\eta]} + \boldsymbol{\beta}_m^{[\lambda]}$ | $\longrightarrow$ | $\tilde{\boldsymbol{\beta}}_{km+1}$ | $\gamma_k^{[\eta]} + \gamma_m^{[\lambda]}$ | $\longrightarrow$ | $\tilde{\gamma}_{km+1}$ |
| $\alpha_{k+1}^{[\eta]} + \alpha_1^{[\lambda]}$ | $\longrightarrow$ | $\tilde{\alpha}_{km+2}$ | $\boldsymbol{\beta}_{k+1}^{[\eta]} + \boldsymbol{\beta}_1^{[\lambda]}$ | $\longrightarrow$ | $\tilde{\boldsymbol{\beta}}_{km+2}$ | $\gamma_{k+1}^{[\eta]} + \gamma_1^{[\lambda]}$ | $\longrightarrow$ | $\tilde{\gamma}_{km+2}$ |
| $\vdots$ | | | $\vdots$ | | | $\vdots$ | | |
| $\alpha_g^{[\eta]} + \alpha_{m-1}^{[\lambda]}$ | $\longrightarrow$ | $\tilde{\alpha}_{gm-1}$ | $\boldsymbol{\beta}_g^{[\eta]} + \boldsymbol{\beta}_{m-1}^{[\lambda]}$ | $\longrightarrow$ | $\tilde{\boldsymbol{\beta}}_{gm-1}$ | $\gamma_g^{[\eta]} + \gamma_{m-1}^{[\lambda]}$ | $\longrightarrow$ | $\tilde{\gamma}_{gm-1}$ |
| $\alpha_g^{[\eta]} + \alpha_m^{[\lambda]}$ | $\longrightarrow$ | $\tilde{\alpha}_{gm}$ | $\boldsymbol{\beta}_g^{[\eta]} + \boldsymbol{\beta}_m^{[\lambda]}$ | $\longrightarrow$ | $\tilde{\boldsymbol{\beta}}_{gm}$ | $\gamma_g^{[\eta]} + \gamma_m^{[\lambda]}$ | $\longrightarrow$ | $\tilde{\gamma}_{gm}$ |

$$
\begin{aligned}
& + \frac{\sum_{j=1}^{m}\gamma_j^{[\lambda]}\exp\left(\alpha_j^{[\lambda]}+\boldsymbol{x}^T\boldsymbol{\beta}_j^{[\lambda]}\right)\sum_{k=1}^{g}\exp\left(\alpha_i^{[\eta]}+\boldsymbol{x}^T\boldsymbol{\beta}_i^{[\eta]}\right)}{\sum_{k=1}^{g}\exp\left(\alpha_k^{[\eta]}+\boldsymbol{x}^T\boldsymbol{\beta}_k^{[\eta]}\right)\sum_{l=1}^{m}\exp\left(\alpha_l^{[\lambda]}+\boldsymbol{x}^T\boldsymbol{\beta}_l^{[\lambda]}\right)} \\
= & \frac{\sum_{i=1}^{g}\sum_{l=1}^{m}\gamma_i^{[\eta]}\exp\left(\left[\alpha_i^{[\eta]}+\alpha_j^{[\lambda]}\right]+\boldsymbol{x}^T\left[\boldsymbol{\beta}_i^{[\eta]}+\boldsymbol{\beta}_j^{[\lambda]}\right]\right)}{\sum_{k=1}^{g}\sum_{l=1}^{m}\exp\left(\left[\alpha_k^{[\eta]}+\alpha_l^{[\lambda]}\right]\boldsymbol{x}^T\left[\boldsymbol{\beta}_k^{[\eta]}+\boldsymbol{\beta}_l^{[\lambda]}\right]\right)} \\
& + \frac{\sum_{i=1}^{g}\sum_{j=1}^{m}\gamma_j^{[\lambda]}\exp\left(\left[\alpha_i^{[\eta]}+\alpha_j^{[\lambda]}\right]+\boldsymbol{x}^T\left[\boldsymbol{\beta}_i^{[\eta]}+\boldsymbol{\beta}_j^{[\lambda]}\right]\right)}{\sum_{k=1}^{g}\sum_{l=1}^{m}\exp\left(\left[\alpha_k^{[\eta]}+\alpha_l^{[\lambda]}\right]+\boldsymbol{x}^T\left[\boldsymbol{\beta}_k^{[\eta]}\boldsymbol{\beta}_l^{[\lambda]}\right]\right)} \\
= & \frac{\sum_{i=1}^{g}\sum_{l=1}^{m}\left(\gamma_i^{[\eta]}+\gamma_j^{[\lambda]}\right)\exp\left(\left[\alpha_i^{[\eta]}+\alpha_j^{[\lambda]}\right]+\boldsymbol{x}^T\left[\boldsymbol{\beta}_i^{[\eta]}+\boldsymbol{\beta}_j^{[\lambda]}\right]\right)}{\sum_{k=1}^{g}\sum_{l=1}^{m}\exp\left(\left[\alpha_k^{[\eta]}+\alpha_l^{[\lambda]}\right]+\boldsymbol{x}^T\left[\boldsymbol{\beta}_k^{[\eta]}\boldsymbol{\beta}_l^{[\lambda]}\right]\right)}.
\end{aligned}
$$

Upon performing the mapping from Table I-B, we can write (6) as

$$
\begin{aligned}
\eta(\boldsymbol{x})\lambda(\boldsymbol{x}) &= \frac{\sum_{k=1}^{gm}\tilde{\gamma}_k\exp\left(\tilde{\alpha}_k+\boldsymbol{x}^T\tilde{\boldsymbol{\beta}}_k\right)}{\sum_{l=1}^{gm}\exp\left(\tilde{\alpha}_l+\boldsymbol{x}^T\tilde{\boldsymbol{\beta}}_l\right)} \\
&= \sum_{k=1}^{gm}\tilde{\gamma}_k\pi_k\left(\boldsymbol{x};\tilde{\boldsymbol{\psi}}_{gm}\right) \\
&= \eta_{gm}\left(\boldsymbol{x};\tilde{\boldsymbol{\omega}}_{gm}\right),
\end{aligned}
$$

where $\tilde{\boldsymbol{\omega}}_{gm}^T=\left(\tilde{\boldsymbol{\psi}}_1,\tilde{\gamma}_1,...,\tilde{\boldsymbol{\psi}}_{gm},\tilde{\gamma}_{gm}\right)$ and $\tilde{\boldsymbol{\psi}}_k^T=\left(\tilde{\alpha}_k,\tilde{\boldsymbol{\beta}}_k\right)$ for $k=1,...,gm$. We obtain the desired result by noting that $\eta_{gm}\left(\boldsymbol{x};\tilde{\boldsymbol{\omega}}_{gm}\right)\in\mathbb{H}$. □

Lemmas 2–6 imply that the class $\mathbb{H}$ satisfies Assumptions (i)–(v) of Theorem 1. Thus, we have the following result.

**Theorem 7.** *The class of functions* $\mathbb{H}$ *is dense in* $\mathbb{C}(\mathbb{X})$.

Since $\mathbb{H}\subset\mathbb{M}$, we obtain the main result of the article in the form of the following corollary.

**Corollary 8.** *The class of functions* $\mathbb{M}$ *is dense in* $\mathbb{C}(\mathbb{X})$.

*Proof.* Since $\mathbb{H}$ is dense in $\mathbb{C}(\mathbb{X})$, by the definition of denseness from Theorem 1, for all $m\in\mathbb{C}(\mathbb{X})$, there exists an $\eta\in\mathbb{H}$ such that $\sup_{\boldsymbol{x}\in\mathbb{X}}|\eta(\boldsymbol{x})-m(\boldsymbol{x})|<\epsilon$, for any $\epsilon>0$. Because $\mathbb{H}\subset\mathbb{M}$, there also exists a $\mu=\eta\in\mathbb{M}$ such that $\sup_{\boldsymbol{x}\in\mathbb{X}}|\mu(\boldsymbol{x})-m(\boldsymbol{x})|<\epsilon$. We get the desired result via the definition of denseness again. □

# 4 Conclusions

In this article, we utilized the Stone-Weierstrass Theorem to prove that the class of MoE mean functions $\mathbb{M}$ is dense in the class of continuous functions $\mathbb{C}(\mathbb{X})$ on the compact domain $\mathbb{X}$. Our result is an alternative to the result of [18], which obtains uniform convergence for target functions that are in a sufficiently-differentiable Sobolev space. Furthermore, we make no restrictions on the shape of the domain $\mathbb{X}$, whereas [18] requires the domain to be a unit hypercube.

Unlike in [18, 8, 7, 13], we do not obtain rates of convergence. Furthermore, our results make no guarantees with respect to the statistical estimability of the MoE mean functions, from data. It is well documented that maximum likelihood (ML) estimation can be used to obtain consistent estimates of the mean function, when the number of components $g$ is known; see [18] for results regarding the ML estimator for MoE models with normal component PDFs, [9] for results regarding the ML estimator when the component PDFs are from the one-parameter exponential family, and [14] for the case of Laplace component PDFs. Results regarding regularized ML estimation of MoE models were obtained in [11]. In [14] and [5], the Bayesian information criterion [15] is demonstrated to be effective for the determination of an unknown $g$.